\relax
\documentclass[letterpaper]{article} 
\usepackage{aaai22}  
\usepackage{times}  
\usepackage{helvet}  
\usepackage{courier}  
\usepackage[hyphens]{url}  
\usepackage{graphicx} 
\usepackage{natbib}  
\usepackage{caption} 
\DeclareCaptionStyle{ruled}{labelfont=normalfont,labelsep=colon,strut=off} 
\usepackage{algorithm}
\usepackage{algorithmic}

\usepackage{microtype}
\usepackage{amsthm}
\usepackage{algorithm}
\usepackage{helvet}
\usepackage{courier}
\usepackage{textcomp}
\usepackage{enumerate}
\usepackage{tabularx}
\usepackage{multirow}
\usepackage{array}
\usepackage{graphics}
\usepackage{graphicx}
\usepackage{subfig}
\usepackage{color}
\usepackage{amsmath}
\usepackage{makecell}
\usepackage{multirow}
\usepackage{amssymb}
\usepackage{comment}
\usepackage{mathrsfs}
\usepackage{extarrows}
\usepackage{booktabs}

\usepackage{newfloat}
\usepackage{listings}
\floatstyle{ruled}
\newfloat{listing}{tb}{lst}{}
\floatname{listing}{Listing}

\title{Constructing Phrase-level Semantic Labels to Form Multi-Grained Supervision \\ for Image-Text Retrieval}

\author {
    Zhihao Fan\textsuperscript{\rm 1},
    Zhongyu Wei\textsuperscript{\rm 1,4}\footnote{Corresponding Author},
    Zejun Li\textsuperscript{\rm 1},
    Siyuan Wang\textsuperscript{\rm 1},\\
    Haijun Shan\textsuperscript{\rm 2},
    Xuanjing Huang\textsuperscript{\rm 1},
    Jianqing Fan\textsuperscript{\rm 3}
}
\affiliations {
    \textsuperscript{\rm 1} Fudan University,
    \textsuperscript{\rm 2} Zhejiang Lab,
    \textsuperscript{\rm 3} Princeton University, \\
    \textsuperscript{\rm 4} Research Institute of Intelligent and Complex Systems, Fudan University, China 
    \{fanzh18,zywei,zejunli20,wangsy18,xjhuang\}@fudan.edu.cn, workingshan@163.com
}

\usepackage{bibentry}

\begin{document}
\maketitle
\begin{abstract}
Existing research for image text retrieval mainly relies on sentence-level supervision to distinguish matched and mismatched sentences for a query image. However, semantic mismatch between an image and sentences usually happens in finer grain, i.e., phrase level. In this paper, we explore to introduce additional phrase-level supervision for the better identification of mismatched units in the text. In practice, multi-grained semantic labels are automatically constructed for a query image in both sentence-level and phrase-level. We construct text scene graphs for the matched sentences and extract entities and triples as the phrase-level labels. In order to integrate both supervision of sentence-level and phrase-level, we propose \textbf{S}emantic \textbf{S}tructure \textbf{A}ware \textbf{M}ultimodal \textbf{T}ransformer (SSAMT) for multi-modal representation learning. Inside the SSAMT, we utilize different kinds of attention mechanisms to enforce interactions of multi-grain semantic units in both sides of vision and language. For the training, we propose multi-scale matching losses from both global and local perspectives, and penalize mismatched phrases. Experimental results on MS-COCO and Flickr30K show the effectiveness of our approach compared to some state-of-the-art models.

\end{abstract}

\section{Introduction}
\begin{figure}[!th]
\centering
\includegraphics[width=0.48\textwidth]{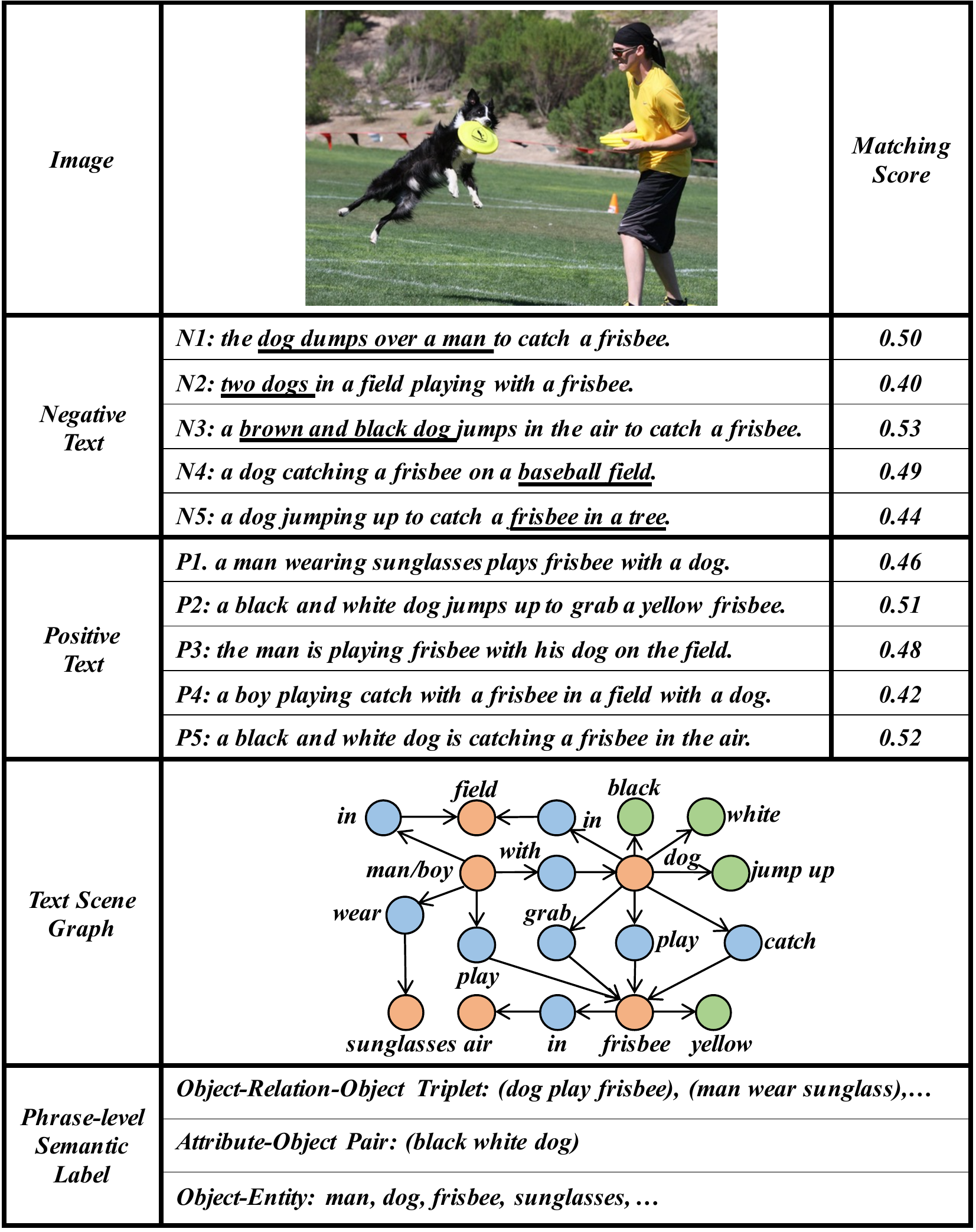}
\centering
\caption{An example of a query image, a group of mismatched sentences, a group of matched sentences and their corresponding text scene graph, and augmented labels in phrase-level. Textual segments with underlines stand for mismatching in phrase-level. The matching scores are produced by VSE++~\cite{faghri2017vse++}.}
\label{intro_example}
\end{figure}
Vision and language are two important aspects of human intelligence to understand the world. To bridge vision and language, researchers pay increasing attention to multi-modal tasks. Image-text retrieval~\cite{NIPS2013_7cce53cf}, one of the fundamental topics, aims to retrieve the matching text (image) for the query image (text). Researchers~\cite{NIPS2013_7cce53cf,kiros2014unifying,you2018end} extract features from an image-text pair to compute a scalar matching score to measure the similarity. The model is optimized via a triplet loss that makes the representations of the positive image-text pair closer than negative ones. Existing research~\cite{faghri2017vse++,lee2018stacked,liu2020graph,wei2020multi} usually relies on sentence-level supervision for cross-modality representation learning. However, the semantic mismatch usually happens in finer grain, i.e., phrase level. 

\begin{figure*}
\centering
\includegraphics[width=0.98\textwidth]{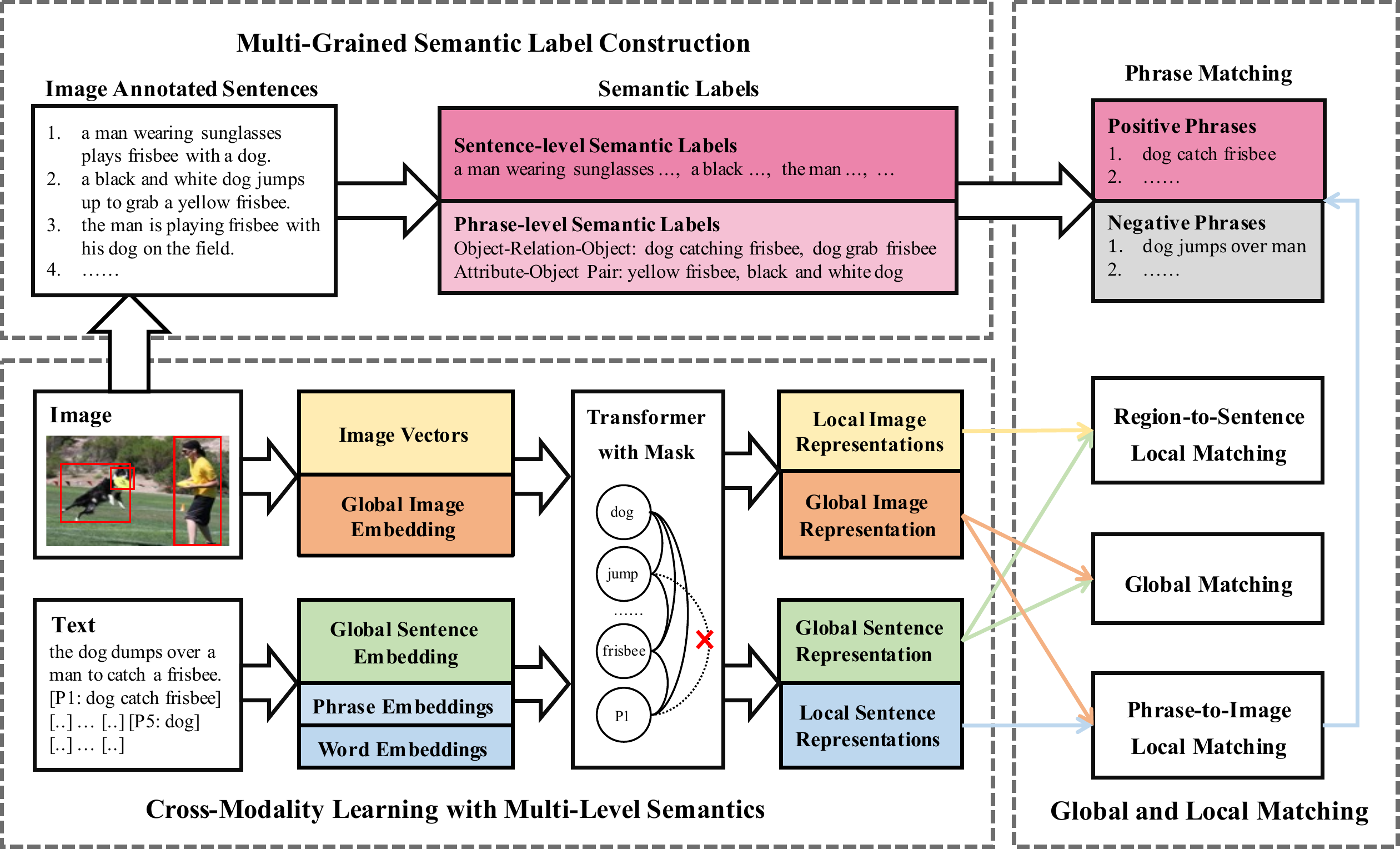}
\centering
\caption{The overall framework of our proposed model Semantic Structure Aware Multimodal Transformer (SSAMT). }
\label{framework_figure}
\end{figure*}


We show an example in Figure~\ref{intro_example}, including a query image, some matched sentences and mismatched ones. In terms of matching scores, the model~\cite{faghri2017vse++} fails to distinguish positive and negative sentences. A closer look at the example shows that mismatched sentences are usually partially irrelevant with phrases of inconsistent semantics (two dogs, baseball filed, etc.). Inspired by this observation, we explore to provide fine-grained supervision in phrase-level for better cross-modality representation learning. In practice, we construct multi-grained semantic labels for a query image of two levels, namely, sentence-level and phrase-level. In sentence-level, we use the whole sentence as the label. In phrase level, we construct the text scene graph of the sentence and extract entities and triples of multiple forms from the graph as labels. Based on these multi-grained semantic labels, we assume the matching model is able to identify fine-grained mismatched semantic units at the same time of distinguishing negative sentences. 

In order to utilize the supervision of both sentence-level and phrase-level for cross-modality representation learning, we propose the Semantic Structure Aware Multimodal Transformer (SSAMT) to model multi-grained semantics in vision and language. In language side, we concatenate the sentence and its phrases as input, while both image and its regions are used in vision side. Mask transformer is used to model semantic units of different granularity for both modalities, besides, novel attention mechanisms are presented for interactions of intra-modality and inter-modality. The model learns representations for both modalities of vision (image and regions) and language (sentence and phrases) in multiple scales (global and local). For optimization, we utilize the global matching and local matching for the similarity measurement of image-text pairs, where global matching computes the matching score of the global representations of the image and text, and local matching measures the similarities from fine-grained perspectives including region-to-text and phrase-to-image. In addition, for the phrases extracted from mismatched sentences, we propose phrase-matching to teach the model to increase scores between the matched image-phrase pairs and decrease those mismatched ones. Experiment results on MS-COCO~\cite{lin2014microsoft} and Flickr30K~\cite{plummer2015flickr30k} show the effectiveness of our model compared to some state-of-the-art approaches. Further analysis reveals that SSAMT is able to provide better interpretability by locating mismatched phrases of negative sentences.

\section{Semantic Structure Aware Multimodal Transformer (SSAMT)}

The overall framework of Semantic Structure Aware Multimodal Transformer is shown in Figure~\ref{framework_figure}. It includes three major components, namely, multi-grained semantic labels construction, cross-modality representation learning with multi-grained semantics and multi-scale matching losses. The multi-grained semantic labels construction is to automatically collect semantic labels from annotated sentences of the query image, the cross-modality representation learning with multi-grained semantics is to capture the semantics of different granularity in both modalities, and multi-scale losses are utilized to measure the similarity between the pair of image and sentence. We take the image $I_{i}$ and text $T_{j}$ as an example to compute their matching score.

\subsection{Multi-Grained Semantic Labels Construction}
\label{configuration_of_semantic_labels}
Each image in vision and language datasets has multiple annotated sentences, for example there are five in MS-COCO~\cite{lin2014microsoft} and Flickr30K~\cite{plummer2015flickr30k}. These sentences describe multi-grained semantics of the image including various objects, relations and scenes, and we propose to utilize them to automatically configure corresponding semantic labels. 

In practice, we adopt the scene graph parser of SPICE~\cite{anderson2016spice} following SGAE\footnote{\url{https://github.com/tylin/coco-caption}}~\cite{yang2019auto} to dig object-relation-object triplets, object-attribute pairs and object entities from upper descriptive sentences. For example in Figure~\ref{framework_figure}, the retrieved phrases include ``dog catching frisbee'' and ``yellow frisbee''. Moreover, tokens of each sentence are also collected as supplementary of the upper phrases. There phrases and tokens are regarded as semantic labels, $L_{i}$, of the image $I_{i}$. Semantic labels provide an opportunity to determine the partially irrelevant components in the sentence $T_{j}$ and we will introduce the details in \S\ref{section_global_local_matching}.

\subsection{Model Input}
\label{model_input_section}




\paragraph{Text Embeddings} 
For the sentence $T_j$, we obtain its tokens and phrases as described in \S\ref{configuration_of_semantic_labels}. To initialize each token, we map it to a dense vector using a standard embedding layer following~\citet{devlin2018bert} which is composed of token embeddings, position embeddings and segment embeddings. For each phrase, we use phrase segment embeddings of three categories, namely, object, attribute and relation as initialization, and think of it as a phrase node that connects to these tokens included in it. We concatenate them with tokens for simultaneous encoding. We do not add context-based embeddings for phrase initialization and introduce the mask mechanism in \S\ref{inter_modality} to encode its context as compensation. At the same time, we set up a global sentence node with dense vector $C^{T}$ as initialization to capture the sentence-level semantics. In summary, our text embeddings have three parts, word embeddings $E^{W}_{j}$, phrase embeddings $E^{P}_{j}$ and a global sentence embedding $C^{T}$.

\paragraph{Image Vectors Initialization}
For the image $I_{i}$, we employ a pre-trained object detector to extract region features, where each $o_{i,k}\in \mathbb{R}^{d_{o}}$ is the mean-pooled convolutional feature for the $k$-th region of $I_{i}$ and $d_{o}$ is the hidden size of the detector. We fix the pre-trained model during training. To fit the hidden size of our encoder, we add a fully-connected layer to project each region feature into the same size and get initial image vectors $E_{j}^{I}$. Following the setting of the global sentence node, we also set up a global image node with $C^{I}$ as initialization to capture the overall semantics of the image.

\subsection{Cross-Modality Modeling with Multi-Grained Semantics}
\label{inter_modality}
To enforce the interaction of multi-grained semantics from both two modalities, we employ inter-modality and intra-modality relationships modeling at the same time, and present the mechanism of mask attention inside the transformer cell to learn the multi-grained semantics with the inherent structure.

\paragraph{Inter-Modality Relationship Modeling}
Inter-modality relationship model aims to set up the interactions across the two modalities. We use the encoder of transformer~\cite{vaswani2017attention} as backbone. In the following equation, we concatenate text embeddings and image vectors in \S\ref{model_input_section} as model input.
\begin{equation}
    \begin{normalsize}
        \begin{gathered}
           H^{0}=\big[C^T,E_j^P,E_j^W;C^I,E_i^I\big] \nonumber
        \end{gathered}        
    \end{normalsize}
    \label{embedding} 
\end{equation}

In the original setting of transformer, there is no different granularity or structure and each element attends to others without constraints. In our case, we have phrase node to capture the semantics of words in the phrase, and modality-dependent global nodes to model the overall semantics for image and text, respectively. These nodes are utilized for multi-grained semantic modeling and heavily depends on the structure. To keep their meaning, we argue to abandon original attention and employ mask attention. In implementation, a masking matrix $M\in\mathbb{R}^{|H^0|\times |H^0|}$ is initialized with all $0$, and we reset values in specific positions with $-\infty$ to meet these three requirements as following.
\begin{enumerate}[(1)]
\setlength{\topsep}{0pt}
\setlength{\itemsep}{0pt}
\setlength{\parsep}{0pt} 
    \item In the vision side, each region node is not visible to
    the global sentence node. 
    \item In the language side, each phrase and token node is not visible to the global image node. 
    \item Each phrase node is not visible to any other words that not included in the phrase itself. 
\end{enumerate}

We add $M$ to the following attention function and utilize it to replace the original one in transformer. We call the new one mask transformer.
\begin{equation}
    \begin{normalsize}
        \begin{gathered}
            \text{attention}(Q,K,V,M)=\text{softmax}\bigg(M+\frac{Q^{T}K}{\sqrt{d_{k}}}\bigg)V~\label{mattn} \nonumber
        \end{gathered}
    \end{normalsize}
\end{equation}

After inter-modality relationship modeling, we get a sequence of outputs shown in the following equation.
\begin{equation}
    \begin{normalsize}
        \begin{aligned}
            H=\big[h_{j}^{{T}},H_{j}^{T};h_{i}^{{I}},H_{i}^{I}\big]\label{encoder_output}\nonumber
        \end{aligned}
    \end{normalsize}
\end{equation}
where $h_{i}^{I}$ and $h_{j}^{T}$ are global representations for image and text corresponding to global nodes $C^{I}$ and $C^{T}$ of image and sentence. $H_{i}^{I}$ are representations for regions and $H_{j}^{T}$ are for phrases and words. They are local representations for the image and sentence, respectively.

\paragraph{Intra-Modality Relationship Modeling} Intra-modality relationship model~\cite{wei2020multi,yang2019context} is employed to separately encode image and text as a supplementary to the inter-modality relationship modeling, where inputs of image and text are $\big[C^I,E_i^I\big]$ and $\big[C^T,E_j^P,E_j^W\big]$, respectively. We take the outputs of $C^{I}$ and $C^{T}$ as intra-modality global representations of the image and text, denoted as $a_{i}^{I}$ and $a_{j}^{T}$.




\begin{table*}
\begin{center}
\resizebox{1.0\textwidth}{!}{
\begin{tabular}{l|ccc|ccc|c|ccc|ccc|c}
\midrule[1.0pt]
&\multicolumn{7}{c}{MS-COCO 1K} &\multicolumn{7}{|c}{Flickr30K}\\
\midrule[1.0pt]
\multirow{2}{*}{Model} 
&\multicolumn{3}{c|}{Image-to-Text} &\multicolumn{3}{c|}{Text-to-Image} & &\multicolumn{3}{c|}{Image-to-Text} &\multicolumn{3}{c}{Text-to-Image} & \\
&R@1 &R@5 &R@10 &R@1 &R@5 &R@10 &RSum &R@1 &R@5 &R@10 &R@1 &R@5 &R@10 &RSum\\
\midrule[1.0pt]
\emph{VSE++} &64.7 &- &95.9 &52.0 &- &92.0 &- &52.9 &79.1 &87.2 &39.6 &69.6 &79.5 &407.9\\
\emph{CAMP} &72.3 &94.8 &98.3 &58.5 &87.9 &95.0 &506.8 &68.1 &89.7 &95.2 &51.5 &77.1 &85.3 &466.9 \\
\emph{SCAN} &72.7 &94.8 &\textbf{98.4} &58.8 &88.4 &94.8 &507.9 &67.4 &90.3 &95.8 &48.6 &77.7 &85.2 &465.0 \\
\emph{SGM} &73.4 &93.8 &97.8 &57.5 &87.3 &94.3 &504.1 &71.8 &91.7 &95.5 &53.5 &79.6 &86.5 &478.6\\
\emph{VSRN} &74.0 &94.3 &97.8 &60.8 &88.4 &94.1 &509.4 &70.4 &89.2 &93.7 &53.0 &77.9 &85.7 &469.9\\
\emph{BFAN} &74.9 &95.2 &- &59.4 &88.4 &- &- &68.1 &91.4 &- &50.8 &78.4 &- &- \\
\emph{MMCA} &74.8 &95.6 &97.7 &61.6 &\textbf{89.8} &95.2 &514.7 &74.2 &\textbf{92.8} &96.4 &54.8 &81.4 &87.8 &487.4\\
\emph{GSMN} &76.1 &95.6 &98.3 &60.4 &88.7 &95.0 &514.1 &71.4 &92.0 &96.1 &53.9 &79.7 &87.1 &480.2 \\
\midrule[1.0pt]
\emph{SSAMT} &\textbf{78.2} &\textbf{95.6} &98.0 &\textbf{62.7} &89.6 &\textbf{95.3} &\textbf{519.4} &\textbf{75.4} &92.6 &\textbf{96.4} &\textbf{54.8} &\textbf{81.5} &\textbf{88.0} &\textbf{488.7}\\
\midrule[1.0pt]
\end{tabular}
}
\end{center}
\caption{Comparison results of the cross-modal retrieval on the MS-COCO 1K and Flickr30K in terms of Recall@K(R@K). The comparative models include VSE++~\cite{faghri2017vse++}, CAMP~\cite{wang2019camp}, SCAN~\cite{lee2018stacked}, SGM~\cite{wang2020cross}, VSRN~\cite{li2019visual}, BFAN~\cite{liu2019focus}, MMCA~\cite{wei2020multi}, GSMN~\cite{liu2020graph}.
} 
\label{OverallPerformance}
\end{table*}



\subsection{Multi-scale Matching Losses}
\label{section_global_local_matching}
Supposing we have a positive image-text pair $(I_i, T_i)$ with a negative image $I_{k}$ and a negative sentence $T_{j}$. We use triplet loss $\text{TriL}_{\alpha}$ to train our model. In $\text{TriL}_{\alpha}(u,V,W)$ as following, $\alpha$ is a scalar to control the distance between the cosine score of $u$ and positive samples $V$ and that of negative samples $W$. The loss is to push $v\in V$ closer to $u$ and push $w\in W$ away from $u$. Based on multi-grained semantic labels, We measure the similarity of these image-text pairs using three kinds of matching scores, including global, local, and phrase matching.
\begin{equation}
\begin{normalsize}
    \begin{aligned}
            &\text{TriL}_{\alpha}\big(u,V,W\big)\\
        =&\mathop{max}\Big(\alpha-\frac{1}{{|V|}}\sum_{v\in V}\mathop{cos}(u,v)+\frac{1}{{|W|}}\sum_{w\in W}\mathop{cos}(u,w),\ 0\Big) \nonumber
        \label{TL}  
    \end{aligned}
\end{normalsize}  
\end{equation}




\paragraph{Global Matching} Intra-modality and inter-modality relationship modeling both produce representations for global representations of image and sentence, then we have $cos(a_{i}^{I},a_{i}^{T})$ and $cos(h_{i}^{I},h_{i}^{T})$ to measure global similarity of the positive image-text pair $(I_i, T_i)$, and so is for the negative pair $(I_i, T_j)$. The corresponding loss is in Eq.~(\ref{L_G_0}).
\begin{equation}
    \begin{normalsize}
        \begin{aligned}
        \mathcal{L}^{G}_{\mathit{0}}&=\text{TriL}_{\alpha_{0}}(a_{i}^{I},a_{i}^{T},a_{j}^{T})+\text{TriL}_{\alpha_{0}}(a_{i}^{T},a_{i}^{I},a_{k}^{I})\\
        \mathcal{L}^{G}_{\mathit{1}}&=\text{TriL}_{\alpha_{1}}(h_{i}^{{I}},h_{i}^{{T}},h_{j}^{{T}})+\text{TriL}_{\alpha_{1}}(h_{i}^{{T}},h_{i}^{{I}},h_{k}^{I}) \\
        \end{aligned}
    \end{normalsize}
    \label{L_G_0}
\end{equation}

\paragraph{Local Matching} We utilize local matching which is based on inter-modality relationship modeling to enhance the fine-grained cross-modal matching. It has two parts. (1) Region-to-Sentence: The matching between each region and the sentence. (2) Phrase-to-Image: The similarity of each phrase (token) and the image. We employ the loss in Eq.~(\ref{L_L_2}) to make local matching scores of the positive image-text pair larger than the negative one.
\begin{equation}
    \begin{normalsize}
        \begin{aligned}
        \mathcal{L}^{L}_{\mathit{2}}&=\text{TriL}_{\alpha_{2}}(h_{i}^{{I}},H_{i}^{T},H_{j}^{T})+\text{TriL}_{\alpha_{2}}(h_{i}^{{T}},H_{i}^{I},H_{k}^{I})\\ 
        \end{aligned}
    \end{normalsize}
    \label{L_L_2} 
\end{equation}

\paragraph{Phrase Matching} 
On the basis of phrase-to-image matching, we employ semantic labels $L_{i}$ to determine mismatched phrases in negative sentences, and decrease the scores of the mismatched ones and increase the scores of matched ones. In detail, for each phrase or token in $T_j$, we determine it as positive if it appears in semantic labels $L_{i}$ or included in some label of $L_i$, otherwise, it is negative. Through the method, we split the token and phrase of $T_j$ into positive ones $H^{T}_{j+}$ and negative ones $H^{T}_{j-}$. We repeat the same process on the negative pair $(I_k, T_i)$ and get $H^{T}_{i+}$ and $H^{T}_{i-}$. Consider that positive parts are keys to separate mismatched image text pair, we propose $\mathcal{L}^{P}_{\mathit{3}}$ in Eq.~(\ref{L_P_3}) to further push away negative parts against positive ones in the negative sentence. It also can be interpreted as the penalty on mismatched parts, which is to guide the matching model to make decisions more grounding on them.
\begin{equation}
    \begin{normalsize}
        \begin{aligned}
        \mathcal{L}^{P}_{\mathit{3}}&=\text{TriL}_{\alpha_{3}}(h_{i}^{{I}},H^{T}_{j+},H^{T}_{j-})+\text{TriL}_{\alpha_{3}}(h_{k}^{{I}},H^{T}_{i+},H^{T}_{i-})\\    
        \end{aligned}
    \end{normalsize}
    \label{L_P_3} 
\end{equation}
Based on these three types of matching methods and corresponding losses, we get the overall loss as Eq.~(\ref{overall_loss}) with hyperparameters $\lambda_{0}$, $\lambda_{1}$, $\lambda_{2}$ and $\lambda_{3}$ to balance these losses.
\begin{equation}
    \begin{normalsize}
        \begin{aligned}
        \mathcal{L}^{\mathit{S}}&=\lambda_{0}\mathcal{L}^{G}_{\mathit{0}}+\lambda_{1}\mathcal{L}^{G}_{\mathit{1}}+\lambda_{2}\mathcal{L}^{L}_{\mathit{2}}+\lambda_{3}\mathcal{L}^{P}_{\mathit{3}}        
        \end{aligned}
    \end{normalsize}
    \label{overall_loss} 
\end{equation}



Previous image-text retrieval models usually take the hardest image (text) from in-batch data as the negative image (text), which requires the matching score computation of all pairwise image-text combinations in batch. This is expensive in inter-modality relationship modeling, thus we sample negative instances through intra-modality matching scores to reduce the computation cost. 

\paragraph{Inference} During inference, we utilize the following $\text{score}(I_{i},T_{j})$ for ranking.
\begin{equation}
    \begin{normalsize}
        \begin{aligned}
            &\text{score}(I_{i},T_{j})=cos(a_{i}^{I},a_{j}^{T})+\mu_{1}cos(h_{i}^{I},h_{j}^{T})\\
            &+\frac{\mu_{2}}{|H_{i}^{T}|}\sum_{h_{j,k}^{T}\in H_{j}^{T}}cos(h_{i}^{I},h_{j,k}^{T})+\frac{\mu_{2}}{|H_{i}^{I}|}\sum _{h_{i,k}^{V}\in H_{i}^{I}}cos(h_{j}^{T},h_{i,k}^{I})
        \label{testing_score_function}  
        \end{aligned}
    \end{normalsize}
\end{equation}
where $\mu_{1}$ and $\mu_{2}$ are hyperparameters.

\section{Experiment}
\begin{table}
\begin{center}
\resizebox{0.47\textwidth}{!}{
\begin{tabular}{l|ccc|ccc}
\midrule[1.0pt]
\multirow{2}{*}{Model}  &\multicolumn{3}{c|}{Image-to-Text} &\multicolumn{3}{c}{Text-to-Image} \\
&R@1 &R@5 &R@10 &R@1 &R@5 &R@10 \\
\midrule[1.0pt]
\emph{CAMP} &50.1 &82.1 &89.7 &39.0 &68.9 &80.2\\
\emph{SCAN} &50.4 &82.2 &90.0 &38.6 &69.3 &80.4\\
\emph{SGM} &50.0 &79.3 &87.9 &35.3 &64.9 &76.5 \\
\emph{MMCA} &54.0 &82.5 &90.7 &38.7 &69.7 &\textbf{80.8}\\
\midrule[1.0pt]
\emph{SSAMT} &\textbf{57.7} &\textbf{84.2} &\textbf{90.8}  &\textbf{40.8} &\textbf{70.5} &80.5 \\
\midrule[1.0pt]
\end{tabular}
}
\end{center}
\caption{Comparison results of the cross-modal retrieval on the MS-COCO in terms of Recall@K(R@K).} 
\label{OverallPerformance5K}
\end{table}

\subsection{Experimental Setup}
\paragraph{Datasets} We evaluate our proposed model on MS-COCO~\cite{lin2014microsoft} and Flickr30K~\cite{plummer2015flickr30k}. Each image of MS-COCO is accompanied with 5 human annotated captions. We split the dataset into training, validation and test sets respectively with $113,287/5,000/5,000$ images following~\cite{karpathy2015deep}. For MS-COCO 1K, the testing set is further divided into 5 splits and the performance reported are the average over the 5 folds of 1K test images~\cite{faghri2017vse++}. 
Flickr30K~\cite{plummer2015flickr30k} consists of 31000 images collected from the Flickr website. Each image contains 5 descriptive sentences. We take the same splits for training, validation and testing sets as in~\citet{karpathy2015deep}, 1000 images for validation and 1000 images for testing, while the rest for training.

\paragraph{Evaluation Metric}
The performance of image text retrieval is evaluated by the standard recall at K (R@K), $K=1,5,10$. It is defined as the fraction of queries for which the correct item belongs to the top-$K$ retried items. In image text retrieval, we can take image or text as query to retrieve matched texts or images, corresponding to the settings of image-to-text and text-to-image. We also take RSum which is the sum of R@1+R@5+R@10 in both image-to-text and text-to-image as an overall metric.

\subsection{Implementation Details}
For the image, we employ Faster-RCNN~\cite{ren2015faster} pre-trained on Visual Genome~\cite{krishna2017visual} to extract region features following BUTD\footnote{\url{https://github.com/peteanderson80/bottom-up-attention}}~\cite{anderson2018bottom}. We prune the vocabulary by dropping words that appear less than five times. Both of our intra-relationship model and inter-relationship model has $2$ layers, the hidden dimension is $1024$, the head of attention is $16$ and the inner dimension of feed-forward network is $2,048$. The number of parameters in our model is $64.7$M. 

Our training has two phases. For all experiments, the dropout rate is $0.3$ and the Adam~\cite{kingma2014adam} with $\beta_{1}=0.9$, $\beta_{2}=0.999$ is taken as the optimizer of our model. The linear-decay learning rate scheduler is employed with 10K update steps, 1K warm-up steps. In MS-COCO, we first train the intra-r model with $\alpha_{0}=0.1$, the maximum instance (\#region+\#token) per batch is $32,768$, the accumulation step is $1$ and the peak learning rate is 2e-4. Then, we fix the intra-modality relationship model to train the inter-modality relationship model with $\alpha_{1}=0.1$, $\alpha_{2}=0.2$, $\alpha_{3}=0.2$ and $\lambda_{1}=1.0$, $\lambda_{2}=1.0$, $\lambda_{3}=0.1$. The maximum instance per batch is $8,192$, the accumulation step is $16$ and the peak learning rate is 6e-4. During testing, $\mu_{1}=0.2$ and $\mu_{2}=0.1$. In Flickr30K, we first train the intra-modality relationship model with $\alpha_{0}=0.05$, the maximum instance per batch is $32,768$, the accumulation step is 1, and the peak learning rate is 1e-4. Then, we fix the intra-modality relationship model to train the inter-modality one with $\alpha_{1}=0.05$, $\alpha_{2}=0.05$, $\alpha_{3}=0.1$ and $\lambda_{1}=1.0$, $\lambda_{2}=0.1$ $\lambda_{3}=0.1$. The maximum instance per batch is $8,192$, the accumulation step is $4$ and the peak learning rate is 8e-4. During testing, $\mu_{1}=0.3$ and $\mu_{2}=0.1$.


\subsection{Overall Performance}
We compare our model with some classic and state-of-the-art approaches, including VSE++~\cite{faghri2017vse++}, CAMP~\cite{wang2019camp}, SCAN~\cite{lee2018stacked}, SGM~\cite{wang2020cross}, VSRN~\cite{li2019visual}, BFAN~\cite{liu2019focus}, MMCA~\cite{wei2020multi}, GSMN~\cite{liu2020graph}. The results on MS-COCO 1K and Flickr30K are presented in Table~\ref{OverallPerformance}, and that on MS-COCO is shown in Table~\ref{OverallPerformance5K}. We can see that our proposed SSAMT outperforms all existing methods, with the
best R@1$=78.2\%$ for image-to-text retrieval and R@1$=62.7\%$ for text-to-image retrieval on MS-COCO 1K. For MS-COCO, the proposed approach maintains the superiority with an improvement of more than $3\%$ on the R@1 of image-to-text retrieval. In Flickr30K, our model achieves the best performance with image-to-text R@1 of $75.4\%$.


\begin{table}
\begin{center}
\resizebox{0.47\textwidth}{!}{
\begin{tabular}{l|cccccc}
\midrule[1.0pt]
&\multicolumn{3}{c|}{Image-to-Text} &\multicolumn{3}{c}{Text-to-Image} \\
 &R@1 &R@5 &R@10 &R@1 &R@5 &R@10 \\
\midrule[1.0pt]
&\multicolumn{6}{c}{mask transformer w/o phrases} \\
\midrule[0.5pt]
\emph{MT+GM} &73.9 &93.7 &97.5 &57.7 &87.0 &94.1 \\
\emph{+LM} &75.5 &94.3 &97.7 &59.8 &87.9 &94.5 \\
\emph{+PM} &76.3 &94.7 &98.0 &61.2 &88.8 &94.8 \\
\midrule[1.0pt]
&\multicolumn{6}{c}{mask transformer w/ phrases} \\
\midrule[0.5pt]
\emph{MT+GM} &74.5 &94.1 &97.3 &58.2 &87.7 &94.4\\
\emph{+LM} &76.8 &95.0 &97.9 &61.4 &88.7 &94.6\\
\emph{+PM} &\textbf{78.2} &\textbf{95.6} &\textbf{98.0} &\textbf{62.7} &\textbf{89.6} &\textbf{95.3}\\ 
\midrule[1.0pt]
\end{tabular}
}
\end{center}
\caption{Ablation study for SSAMT on MS-COCO 1K. There are two groups of settings, namely, mask transformer w/o phrases and mask transformer w/ phrases. In each group, we take mask transformer with global matching loss as baseline. Components are added on top of the previous setting one by one from the first row to the bottom one.} 
\label{ablation_study}
\end{table}

\subsection{Ablation Study}
\label{section_ablation_study}
We perform ablation study on MS-COCO 1K to explore the effectiveness of phrase-level labels and different matching scores. We compare two groups of settings by controlling the usage of phrase-level labels, namely, mask transformer w/o phrases and w/ phrases. Under each group of settings, we take mask transformer (MT)  with global matching loss (GM) as baseline. On top of it, we add local matching loss (LM) and phrase matching loss (PM) in sequence to justify their influences. Experiment results are shown in Table~\ref{ablation_study}. We can see that performance increases as components are gradually added. Moreover, these models based on mask transformer w/ phrases perform better than their count-parts without phrases. These facts demonstrate the effectiveness of different components of SSAMT. 


\section{Further Analysis}
In this section, we dive into SSAMT to further analyze the characteristics of semantic labels.
\begin{figure*}[htbp]
\centering
\includegraphics[width=\textwidth]{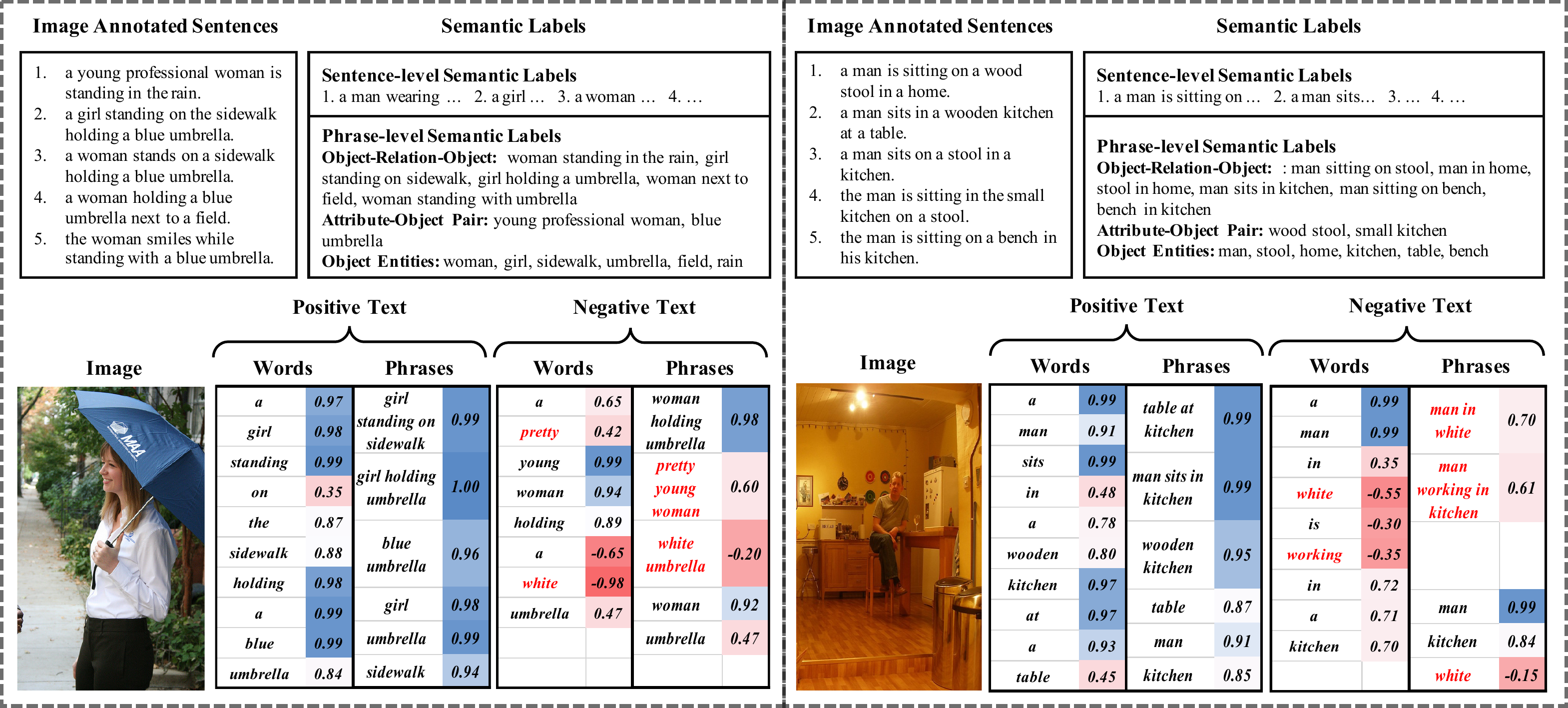}
\centering
\caption{Two examples of SSAMT. Red words in negative text mean they are negative that does not appear in semantic labels. Score on the right of each phrase or word is the corresponding phrase-to-image local matching score. Color of scores from blue to red denotes that phrases or words are more and more irrelevant to the image.}
\label{case_study_example}
\end{figure*}

\subsection{Influence of Phrase-level Labels on Inference Power of Various Matching Scores}
Multi-grained semantic labels are used in computing the multi-scale scores for a given image text pair as Eq.~(\ref{L_G_0}$\sim$\ref{L_P_3}). Experiments have shown their effectiveness on improving the overall performance in image-text retrieval. We would like to further explore contributions of different matching scores in the inference process. In addition to the full version of SSAMT, we also train SSAMT without phrase matching loss (denoted as SSAMT w/o PM) as comparison. SSAMT w/o PM has the same architecture as SSAMT and is able to compute phrase matching score during inference. In the experiment of image-to-text (i2t) setting, for each positive image text pair in the test set of MS-COCO~\cite{lin2014microsoft}, we sample another sentence to form a negative pair. So is text-to-image (t2i). We compute the mean of classification accuracy across the test set.

Experiment results are presented in Table~\ref{loss_validation}. We use one of three matching scores (global, local and phrase) to make the decision of classification. Two findings standout. (1) The performance of SSAMT w/o PM drops significantly when it computes phrase matching to determine the irrelevant parts of the negative sentence. This indicates that if there is no explicit supervision to guide the model to be grounding on the truly irrelevant parts during training, it hardly unsupervised learns the grounding. (2) SSAMT outperforms SSAMT w/o PM in all six metrics. This demonstrates that, through guiding the model to be more grounding on the truly irrelevant parts, phrase-level labels contribute to all three matching during the inference. 


\begin{table}
\begin{center}
\resizebox{0.42\textwidth}{!}{
\begin{tabular}{c|cc}
\midrule[1.0pt]

Matching Type &\text{\quad SSAMT\quad } &SSAMT w/o PM\\
\midrule[1.0pt]
global (i2t) &76.9\% &75.2\% \\
global (t2i) &83.9\% &82.6\% \\
\midrule[0.5pt]
local (i2t) &77.2\% &75.4\% \\
local (t2i) &84.2\% &82.6\% \\
\midrule[0.5pt]
phrase (i2t) &97.1\% &30.0\% \\
phrase (t2i) &97.2\% &28.4\% \\
\midrule[1.0pt]
\end{tabular}
}
\end{center}
\caption{Accuracy of global matching, local matching and phrase matching in Eq.~(\ref{L_G_0}$\sim$\ref{L_P_3}) with respect to SSAMT and SSAMT w/o PM. i2t and t2i mean image-to-text and text-to-image, respectively.} 
\label{loss_validation}
\end{table}

\begin{table}
\begin{center}
\resizebox{0.45\textwidth}{!}{
\begin{tabular}{l|cccc}
\midrule[1.0pt]
Model &100\% &75\% &50\% &25\% \\
\midrule[1.0pt]
SSAMT w/o PM\&LM &483.3 &462.4 &436.5 &372.5 \\
SSAMT w/o PM &485.1 &464.6 &437.7 &373.4\\
SSAMT &488.7 &471.2 &446.4 &383.8 \\
\midrule[1.0pt]
\end{tabular}
}
\end{center}
\caption{RSum of different models with different sizes of training data.} 
\label{data_efficiency_result}
\end{table}

\subsection{Influence of Phrase-level Labels on the Improvement of Training Efficiency}
In the application of semantic labels, we exploit more supervision signals for each query image in phrase level. We would like to see the influence of these phrase-level labels on the training efficiency. We compare three versions of SSAMT, namely, SSAMT, SSAMT w/o PM and SSAMT w/o PM\&LM, with different sets of matching losses respectively. In practice, we randomly pick out 75\%, 50\% and 25\% data from Flickr30K~\cite{plummer2015flickr30k}, which respectively contain $21,750$, $14,500$ and $7,250$ images and $108,750$, $72,500$ and $36,250$ image-text pairs. We present RSum score of three models on test set of Flickr30K for evaluation. 

Experiment results are shown in Table~\ref{data_efficiency_result}. We find that RSum of all three models decreases as dataset size decreases from $100\%$ to $25\%$. The performance gain of SSAMT w/o PM over SSAMT w/o PM\&LM gets smaller as the dataset size decreases, which ranges from 1.8 to 0.9. However, the gain of SSAMT over SSAMT w/o PM gets larger, which ranges from 3.6 to 10.4. This demonstrates that phrase-level labels can improve the training efficiency, and the improvement is more significant with a small dataset scale.

When the size of training dataset reduces from $29,000$ ($100\%$) to $7,250$ ($25\%$), the size of vocabulary changes from $9,568$ to $8,608$, of which the reduction is smaller than the dataset. It has less impact on phrase matching. This means that the supervision of phrase matching does not decrease that rapidly. Thus, semantic labels are more beneficial for image-text retrieval under the condition of small data amount.

\subsection{Case Study}
\label{case_study}
We show two examples in Figure~\ref{case_study_example}.
For each image, we show its annotated sentences on top of it. Based on these sentences, we follow the instruction in \S\ref{configuration_of_semantic_labels} to automatically construct corresponding multi-grained semantic labels, and list them on the right of these image annotated sentences. We show two texts that one is positive (matched) and the other is negative(mismatched) on the right of each image. Each of these texts is separated into words (left) and phrases (right), and they are accompanied with corresponding phrase-to-image local matching scores. We use blue and red to highlight positive and negative scores respectively, the darker the higher (lower in negative). We observe that phrase scores of the positive sentence are large, and those of negative one are relatively small, especially those mismatched parts, such as ``white umbrella", ``man in white" and ``man working in kitchen". These facts verify that SSAMT can produce more faithful results.



\section{Related Work}

Most works in image-text retrieval focus on capturing the cross-modal semantic association by better feature extraction and cross-modality interaction.
\citet{nam2017dual} and~\citet{ji2019saliency} represent the image by semantics gathered from block-based attention, or by region-level.~\citet{lee2018stacked},~\citet{wang2019camp},~\citet{wang2019position},~\citet{li2019visual},~\citet{wang2020cross} and~\citet{wei2020multi} detect objects in images by pre-trained Faster R-CNN~\cite{ren2015faster} following the bottom-up manner proposed by \citet{anderson2018bottom}. 
For text processing,~\citet{klein2015associating} explore to use Fisher Vectors as discriminative representations. Language models like Skip-Gram are employed to extract word representations~\cite{frome2013devise}, and text segments are generally encoded by recurrent neural network (RNN)~\cite{kiros2014unifying, faghri2017vse++, chen2020expressing}. Besides,~\citet{yu2019deep},~\citet{wu2019learning} and~\citet{wei2020multi} use the mechanism of self-attention for enforce the relationship modeling in sentence. On top of fine-grained features, a line of research propose to model inter-modality interaction, such as dual attention network~\cite{nam2017dual}, stacked cross attention~\cite{lee2018stacked,liu2019focus,hu2019multi}, graph structure attention~\cite{li2019visual,wang2020cross,liu2020graph}, and the multi-modal transformer modeling~\cite{wei2020multi}. UNITER~\cite{chen2020uniter} and Unicoder~\cite{li2020unicoder} follow BERT~\cite{devlin2018bert} to pre-train the vision-language transformer model on the large-scale image-text datasets, and finetune in image-text retrieval. 
Compare with these transformer-based models, our model expands the transformer input with phrases through introducing phrase nodes and keep their local structure with words during encoding for better multi-grained semantic modeling. 
More important, previous works focus on finding better models for better cross-modality representation learning, and supervision is always the sentence-level triplet loss. In our works, we provides the fine-grained supervision in phrase-level instead of only a sentence-level matching (mismatching) signal, and the supervision guides the model to discriminate mismatched sentences with more grounding on irrelevant local parts. This method not only helps the model to achieve better retrieval performance, but also is more interpretable and faithful.

\section{Conclusion}
\label{SectionConclusion}
In this paper, to make full use of negative sentences in both phrase-level and sentence-level, we explore to build multi-grained semantic labels, in which phrase-level ones are automatically constructed through extracting phrases of object entities, object-attribute pairs and object-relation-object triplets from images annotated sentences. We concatenate the sentence and its phrases in language side, while image and its regions in vision side, then present mask transformer for jointly cross-modality modeling with multi-grained semantics. We have multi-scale matching losses to capture the image-to-text matching and region-to-sentence/phrase-to-image matching. Based on the phrase-to-image matching, we utilize the semantic labels to determine the non-correspondence between phrases and image, and adjust scores between the image-phrase pairs.
Experiment results show the effectiveness of our model on MS-COCO and Flickr30K. Further analysis reveals that semantic labels improve the efficiency of data exploitation and guide the model to discriminate mismatching sentences with more grounding on mismatched parts.

\bibliography{aaai22}

\end{document}